\documentclass[letterpaper, 10 pt, conference]{ieeeconf}  
\IEEEoverridecommandlockouts                                                                                   
\usepackage{graphics} 
\usepackage{epsfig}
\usepackage{mathptmx} 
\usepackage{times} 
\usepackage{amsmath,amsfonts}

\title{\LARGE \bf
Optimal tool path planning for 3D printing with spatio-temporal and thermal constraints }
\author{Zahra Rahimi Afzal$^{1}$, Pavana Prabhakar$^{2}$, Pavithra Prabhakar$^{3}$
  \thanks{$^{1}$Zahra Rahimi Afzal is a graduate student in the Department of Computer Science at the Kansas State University, USA
        {\tt\small zrahimi@ksu.edu}}%
\thanks{$^{2}$Pavana Prabhakar is an assistant professor in the Department of Civil Engineering at the University of Wisconsin, Madison, USA
       {\tt\small pprabhakar4@wisc.edu}}%
\thanks{$^{3}$Pavithra Prabhakar is an associate professor in the Department of Computer Science at the Kansas State University, USA
       {\tt\small pprabhakar@ksu.edu}}%
}

\begin{document}
\maketitle
\thispagestyle{empty}
\pagestyle{empty}

\newcommand{\T}{\ensuremath{\textit{T}}}
\renewcommand{\S}{\ensuremath{\textit{S}}}
\renewcommand{\P}{\ensuremath{\textit{P}}}
\newcommand{\norm}[1]{\ensuremath{|#1|}}
\newcommand{\cost}{\ensuremath{\textit{Cost}}}
\newcommand{\reals}{\ensuremath{\mathbb{R}}}
\renewcommand{\int}{\ensuremath{\mathbb{Z}}}
\newcommand{\enc}{\ensuremath{\textit{Enc}}}
\renewcommand{\t}[3]{\ensuremath{T_{#1,#2}^{#3}}}
\newcommand{\p}[3]{\ensuremath{p_{#1,#2}^{#3}}}
\renewcommand{\u}[3]{\ensuremath{u_{#1,#2}^{#3}}}

\begin{abstract}
In this paper, we address the problem of synthesizing optimal path
plans in $2D$ subject to spatio-temporal and thermal constraints.
Our solution consists of reducing the path planning problem to a Mixed
Integer Linear Programming (MILP) problem.
The challenge is in encoding the ``implication'' constraints in the
path planning problem using only conjunctions that are permitted by
the MILP formulation.
Our experimental analysis using an implementation of the encoding in a
Python toolbox demonstrates the feasibility of our approach in
generating the optimal plans.
\end{abstract}

\section{INTRODUCTION}
Path Planning~\cite{lavalle,latombe} is a classical problem in
robotics that has been extensively investigated.
In this paper, we focus on $2D$ path planning in the presence of
spatio-temporal and thermal constraints to optimize for superior
structural properties of printed $3D$ structures in the context of
additive manufacturing.

Additive manufacturing or the $3D$ printing procedure broadly consists
of placing material in $3$-dimensional space with the help of
computer-controlled material delivery systems,
and fusing new material onto either a printing platform or previously
printed material.
Few common additive manufacturing methods include Fused Filament
Fabrication (FFF), Selective Laser Sintering (SLS) and
Stereolithography (SLA).
In this paper, we focus on FFF process, which is widely used due to
its ease of customization and cost effectiveness.
Broadly, the $3D$ printing procedure takes a $3D$ CAD model of a structure
as input, and generates $2D$ projections of the models at different
layer heights using a software called a ``slicer''.
The sequence of multiple $2$-dimensional layers are then printed to
create a $3$-dimensional part.
To print a particular layer, the printer uses $3$ computer-controlled
motors, one for each axis, to precisely position a material extruder
in $3D$ space which melts a filament in a heating chamber and extrudes
(draws) onto the print bed or onto previously printed parts.
It has been observed that structural properties of the printed parts
are affected by the temperature profile during the printing~\cite{prabhakar18}.

In this paper, we focus on the $2D$ tool path planning, and our
objective is to ensure that the printing pattern is covered by the
printer in the least time, while at the same time maintaining the
temperature between certain bounds.
In particular, this requires us to consider the spatio-temporal heat
evolution model in the planning problem.
We consider a discrete setting, wherein the $2D$ print bed is divided
into a finite number of cells, and the print pattern is specified by a
subset of these cells.
The problem is to synthesize a plan for the traversal of the
printhead/nozzle with the constraint that the nozzle can move one unit
along the $x$-axis or the $y$-axis in each time step.
We also simplify/discretize the heat equation, and include that in our
planning model.

Our broad approach to synthesizing the plan is to encode the optimal path
planning problem into a Mixed Integer Linear Programming (MILP)
problem, and use an off-the-shelf solver to obtain the optimal
assignment to the variables from which a plan can be extracted.
The challenge arises from the fact that a straightforward encoding of
the constraints requires using ``implications'' while MILP only
supports conjunctions of linear constraints.
Hence, we need to choose appropriate variables and define appropriate
linear constraints that can encode these ``implications''.
We have implemented the encoding and the extraction algorithms in a
Python toolbox.
We illustrate our procedure on an example, as well as report our
experimental analysis which demonstrates the feasibility of the
approach to synthesize optimal plans for spatio-temporal and thermal
constraints.

The novelty of our work lies in the fact that while $2D$ path planning
has been investigated extensively, it has not been studied in the
context of spatio-temporal and thermal constraints that need to be
considered to optimize for superior structural properties.

\subsection{Related Work}
Coverage problem consists of generating a plan to cover a given area,
which is closely related to printing problem, that requires covering
a print region.
Path planning problem for coverage have been investigated in
~\cite{galceran2013survey,yang2004neural,choset2001coverage}.  
Optimal path planning has been extensively investigated in the context
of minimizing the total distance
travelled~\cite{stentz1994optimal,yang2002optimal,lepetivc2003time,mcgee2005optimal,
  mcgee2007optimal,wu2000time,bhattacharya2007voronoi,smith2011optimal,prabhakar-cdc-17}.  

In the context of tool path planning for $3D$ printers,
two classes of algorithms have been explored, namely, those that
provide an optimal path to visit a set of points in a two dimensional
plane, and those that traverse a set of print segments such that the
total distance travelled along non-print segments is minimized.
Several strategies for path generation have been explored that do not
necessarily consider optimality including Raster
~\cite{dunlavey1983efficient}, ZigZag ~\cite{park2000tool}, Contour
~\cite{farouki1995path}, Spiral ~\cite{wang2002metric},  and Fractal space
curves ~\cite{kulkarni2000review}. 
Optimal path planning for the first class of problems has been
investigated based on modifications of the asymmetric travelling salesman
problem~\cite{wah2002tool}. 
The second class of problems have been investigated for different
velocity profiles including constant velocity
profiles~\cite{dreifus2017path} and triangular/trapezoidal velocity
profiles
~\cite{fok2017refinement,ganganath2016trajectory,fok2016relaxation,fok20163d}.  
Further, $3D$ path planning algorithms based on Eulerian tour checking algorithm
known as Hierholzer’s algorithm~\cite{dreifus2017path},
tour construction algorithm known as Frederickson's algorithm  
~\cite{fok2017refinement,frederickson1979approximation}, 
Christofides algorithm~\cite{fok2017refinement}, and 
greedy $2$-Opt and greedy annealing
algorithm~\cite{lechowicz2016path}, have been considered.  
While several of these algorithms focus on optimality of the $2D$/$3D$
  path plans with respect to distance travelled, none of them optimize
  the plans for superior structural properties based on thermal
  constraints, which is the focus of this paper.
\section{Preliminaries}
Let $\reals$ denote the set of real numbers and $\int$ the set of
integers.
Let $[n]$ denote the set $\{0, \cdots, n-1\}$.
Given a sequence $\rho = a_0 a_1 a_2 \cdots a_k$, $\rho(i)$ will denote
the $i$-th element, namely, $a_i$, and $k$ will be the length of
$\rho$.
Given an $m \times n$ matrix $P$, we use $P(i, j)$ to denote the
$j$-th element of the $i$-th row of $P$; the row numbers are $0, 1,
\cdots, m-1$ and the column numbers are $0, 1, \cdots, n-1$.
\section{$2D$ path planning problem}
In this section, we discuss the optimal $2D$ path planning problem
under spatio-temporal and thermal constraints towards achieving
superior structural properties, and formalize a discrete version of
the problem. 

\subsection{An overview}
\begin{figure}[thpb]
      \centering
      \includegraphics[scale=.3]{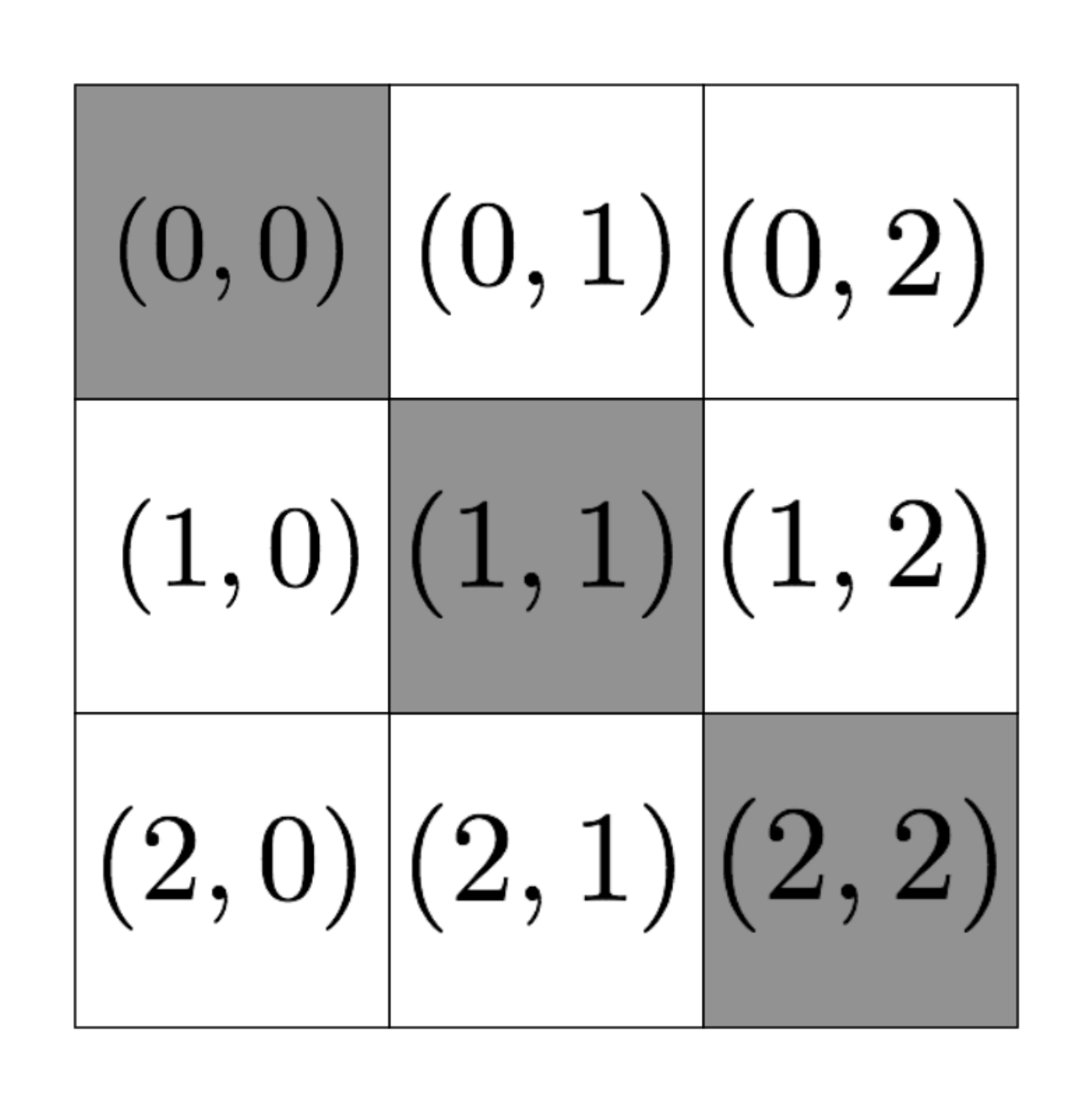}
      \caption{A $3 \times 3$ bed with print region (gray)}
      \label{fig:scenario}
   \end{figure}

We focus on a $2D$ path planning problem that corresponds to printing
a ``slice'' of a given $3D$ object.
The print head can move in the $x$ and $y$ directions, and can extrude
the print material during the movement.
While the print head moves in continuous fragments, we consider
a discrete version of the problem as a simplified first step to
incorporate the spatio-temporal and thermal constraints required to
print objects with superior structural properties.
Hence, we divide the two dimensional plan into an $m \times n$ grid,
and specify a subset of these cells as the print region as shown in
Figure \ref{fig:scenario}.
It shows a $3 \times 3$ grid partitioning the print bed (or a $2D$
layer) into $9$ cells, where the diagonal cells constitute the print pattern.
We allow the print head to move to the adjacent cell either in the
$x$-direction or in the $y$-direction in one time step, but not
simultaneously in both directions.

The nozzle extrudes the material by heating the filament, hence, the
print bed's temperature increases at the point of extrusion of the
material.
The heat evolves in a spatio-temporal manner according to the heat
equation, given by the following partial differential equation (PDE):
\begin{equation}
  \label{eqn:pde}
\frac{\partial{T}}{\partial{t}} = \alpha (
\frac{\partial{^2T}}{\partial{x}^2} +
\frac{\partial{^2T}}{\partial{y}^2} ) + u(x, y, t),
\end{equation}
where $T(x, y, t)$ is the temperature at the position $(x, y)$ at time
$t$, $\alpha$ is the thermal diffusivity, and $u(x, y, t)$ is the
temperature applied at time $t$ to position $(x, y)$, which in our case
will be the temperature of the extruded material.
We consider a simple discrete model of the heat equation, wherein we
replace the second derivatives $\frac{\partial{^2T}}{\partial{x}^2}$
and $\frac{\partial{^2T}}{\partial{y}^2}$ by their first derivatives
$\frac{\partial{T}}{\partial{x}}$ and
$\frac{\partial{T}}{\partial{y}}$, respectively.
Further, we discretize $\frac{\partial{T}}{\partial{x}}$ and
$\frac{\partial{T}}{\partial{y}}$.
Note that $\frac{\partial{T}}{\partial{x}}$ can be approximated as
$\frac{T(x + \Delta, y, t) - T(x, y, t)}{\Delta}$, for a given
$\Delta$.
By taking $\Delta$ to be $1$ (which corresponds to the next
cell along the $x$-axis) we obtain $T(x + 1, y, t) -
T(x, y, t)$.
Similarly, we can approximate the slope from the left, that is,
take $\Delta = -1$, and we obtain  $T(x, y, t) -
T(x-1, y, t)$.
We approximate the partial derivative by the average of the
approximation from the left and the right and obtain  $[T(x + 1, y, t) -
T(x, y, t) + T(x, y, t) - T(x-1, y, t)]/2 = [T(x + 1, y, t) - T(x-1,
y, t)]/2$.
Similarly, we approximate the partial derivative with respect to $y$ by
$[T(x, y + 1, t) - T(x, y-1, t)]/2$.
Hence, we obtain the following difference equation approximating the
heat evolution:

\[T(x, y, t + 1) - T(x, y, t) = \alpha [ T(x + 1, y, t) - T(x-1,
  y, t) +\]
\begin{equation}
  \label{eqn:heat}
  T(x, y + 1, t) - T(x, y-1, t) ]/2 + u(x, y, t)
  \end{equation}

\subsection{Problem Formulation}
We formalize the optimal tool path planning problem.
Formally, the problem is specified using a printing scenario $\S = (m,
n, \P, \T, \T_l, \T_u, \alpha, \T_h)$, where:
\begin{itemize}
\item
  the bed is divided into an $m \times n$ grid of cells;
\item
  $\P$ is an $m \times n$ matrix that specifies the printing pattern, that is,
  $\P(i, j) = 1$ implies that the cell $(i, j)$ needs to be printed, and $\P(i, j) = 0$ implies that cell
  $(i, j)$ does not need to be printed;
\item
  $\T$ is an $m \times n$ matrix that specifies the initial temperature
  profile of the bed, that is, $\T(i, j)$ is a real number that
  specifies the temperature of the cell $(i, j)$ at the beginning of
  printing, that is, at time $0$;
\item
  $\T_l$ and $\T_u$ provide lower and upper bounds on the temperature
  of the bed during the printing; and
\item
$\alpha$ and $\T_h$ are parameters for the temperature evolution
model. 
\end{itemize}

The objective is to find a plan such that the pattern is printed, the
temperature constraints are maintained and the time to print is
minimized.
Two cells positions $(i, j)$ and $(i', j')$ are adjacent if either
$i=i'$ and $\norm{j - j'} = 1$ or $j = j'$ and $\norm{i - i'} = 1$.
A \emph{plan} is a sequence of adjacent cells, that is, $\rho =
\{(x_k,y_k)\}_{k \in [d]}$ such that $(x_{k+1}, y_{k+1})$ is an adjacent
cell of $(x_k, y_k)$, for all $k$.
We refer to $(x_k, y_k)$ as the $k$-th cell visited by the plan.
Let $\cost(\rho)$ denote the distance travelled along $\rho$, that is,
the length of $\rho$.
An \emph{annotated plan} is a pair $(\rho, \{u_k\}_{k  \in [d]},
\{\T_k\}_{k \in [d]})$, where $\rho$ is a plan of length $d$, $u_k
\in \{0, 1\}$ indicates if the nozzle is printing at the $k$-th time
instance in the plan, and $\T_k(i, j)$ specifies the temperature of cell $(i,
j)$ at time $k$.
Our objective is to find an annotated path with minimum cost, such
that the following conditions hold:
\begin{itemize}
\item
{\bf Cover the pattern}: For every $(i, j)$, if $P(i, j) = 1$, then there
is exactly one $k \in \{1, \cdots, d\}$ such that $\rho(k) = (i, j)$
and $b_k = 1$; and if $P(i, j) = 0$, then there
is no $k \in \{1, \cdots, d\}$ such that $\rho(k) = (i, j)$
and $b_k = 1$.
\item
{\bf Temperature evolution constraints:} The temperature profile at
time $k+1$ is related to that at time $k$ according to Equation
(\ref{eqn:heat}). For all $(i, j)$, $\T_0(i, j) = \T(i, j)$, and 
\[
  \T_{k+1}(i, j)= \T_k(i, j)+ \alpha( [\T_k(i+1, j) - \T_k(i-1, j)]/2
\]
\begin{equation}
  + [\T_k(i, j+1) - \T_k(i, j-1)]/2) + T_h u_k b_k,
  \end{equation}
  where $b_k = 1$ if $\rho(k) = (i, j)$, and $0$, otherwise.
  Note that if $(i, j)$ corresponds to a boundary cell, then some of
  the indices on the right are outside the range $[m] \times [n]$, we
  take those $\T_k(i, j)$ to be $0$ (or equivalently, eliminate those
  terms from the equation).
\item
  {\bf Temperature bound constraints:}
  The temperature always remains within the bounds $\T_l$ and $\T_u$.
  For all $i \in [m], j \in [n], k \in [d]$,
  \[\T_l \leq \T_k(i, j) \leq \T_u\]
\end{itemize}
We refer to an annotated path that satisfies the above conditions as \emph{valid}.
\section{Plan Synthesis using MILP}
In this section, we provide a method to synthesize a plan to print a
given pattern with a minimum length plan while satisfying the temperature
constraints.
Our broad approach is to encode the problem as a Mixed Integer Linear
Programming (MILP) problem and to extract a plan from a satisfying
valuation of the constraints.
The challenge arises from the fact that a natural way to encode our
problem involves implications, however, MILP only allows conjunctions,
where as implication requires disjunction.
Hence, we need to use a slightly involved encoding that avoids the
implications.
First, we provide an overview of MILP, and then we present our
encoding.

\subsection{MILP}
Mixed Integer Linear Programming (MILP) refer to an optimization
problem with linear objectives and linear constraints where the
variables consist of a mixture of real and integer variables.
Let $x = (x_0, \cdots, x_{n-1})$ be a tuple of variables, where $I
\subseteq [n]$ is a set of indices which correspond to integer
variables, with the rest being real variables.
The general form of MILP is given by:

\begin{equation*}
\min c^Tx
\end{equation*}
\begin{equation*}
  Ax \leq b, \ x \in \reals^n, \ x_i \in \int, \ \forall i \in I
\end{equation*}
  
Here $c^Tx$ represents a linear objective function, $Ax \leq b$
represents a finite number of linear constraints over the variables
$x$, where the constants $c$, $A$ and $b$ are rationals.
The problem is to find a real valuation for all the variables $x$,
with the additional restriction that the variables with indices from
$I$ have integer values, such that the constraint $Ax \leq b$ is
satisfied and the value of $c^Tx$ is minimized.
While the MILP problem is NP-complete, there are efficient tools such
as Gurobi~\cite{gurobi} and CPLEX~\cite{cplex} that can be used to solve the problems.

\subsection{Encoding}
We present an encoding of the 3D tool path planning problem into an
MILP problem.
Given a printing scenario  $\S = (m, n, \P, \T, \T_l, \T_u, \alpha, \T_h)$
and a time bound $d$, we generate a MILP problem $\enc(\S, d)$
such that the optimal value of $\enc(\S, d)$ corresponds to a shortest
length annotated plan.

$\enc(\S, d)$ uses the following variables:
For every $i \in [m], j \in [n], k \in [d]$,
\begin{itemize}
\item a real-valued variable $\t{i}{j}{k}$ that represents the temperature of the $(i, j)$-th cell in the plan at time $k$.
\item an integer (boolean) variable $\u{i}{j}{k}$ that takes values
  from $\{0, 1\}$, and a value of $1$ means that the nozzle is
  printing at $(i,j)$-th cell at time $k$ and $0$ means that the
  nozzle is not printing at the $(i, j)$-th position at time $k$.
\item an integer (boolean) variable $\p{i}{j}{k}$ that takes values
  from $\{0, 1\}$, and a value of $1$ means that the nozzle is
  positioned at the $(i,j)$-th cell at time $k$ and $0$ means that the
  nozzle is not positioned at the $(i, j)$-th cell at time $k$.
\item an integer variable $m$ when is minimized captures the index in the
  plan corresponding to the last printed cell.
\end{itemize}

$\enc(\S, d)$ consists of the following constraints, for every $i \in
[m], j \in [n], k \in [d]$: 
\begin{itemize}
\item [C1]
The following constraint encodes the temperature evolution; in
particular, it captures the relationship between temperature at time
$k+1$  and at time $k$ as a result of the application of input:

\[\t{i}{j}{k+1} = \t{i}{j}{k} + \alpha( [\t{i+1}{j}{k} - \t{i-1}{j}{k}]/2\]
\[+ [\t{i}{j+1}{k} - \t{i}{j-1}{k}]/2) + \T_h \u{i}{j}{k}\]

The constraint will change accordingly for the boundary cells, in
that, the terms corresponding to the indices that are out of the range
will be eliminated. 

\item [C2]
The following constraints specify that each of the print cells in the
pattern $\P$ is printed at exactly one of the times between $1$ and
$d$.
More precisely, for every $(i, j)$ in the print region, that is, $P(i,
j) = 1$, the nozzle will print at some time between $0$ and $d$, and
it is expressed as: 
$$
\sum_{k \in [d]} \u{i}{j}{k} = 1 
$$
For every $(i, j)$ not in the print region, that is, $P(i,
j) = 0$, the nozzle will not print at any time between $0$ and $d$, and
it is expressed as: 
$$
\sum_{k \in [d]} \u{i}{j}{k} = 0
$$

\item [C3]
We need to encode that if the cell $(i, j)$ is being printed at time $k$, then the
nozzle is also positioned at cell $(i, j)$ at time $k$.
That is, if $\u{i}{j}{k} =1$, then $\p{i}{j}{k} = 1$.
Since, we do not have implications in MILP, we capture it using the
following constraint. 
$$
\u{i}{j}{k} \leq \p{i}{j}{k}
$$
Note that if $\u{i}{j}{k} =1$, it implies that
$\p{i}{j}{k} = 1$.
On the other hand,  if $\u{i}{j}{k} =0$, it specifies a trivial
constraint on $\p{i}{j}{k}$, that is, $\p{i}{j}{k} \geq 0$.

\item [C4]
  Next, we add constraints corresponding to the plan.
  At every time instance $k$, the nozzle is in exactly one
  position. For every $k \in [d]$, 
$$
\sum_{(i, j) \in [m] \times [n]} \p{i}{j}{k} = 1
$$

\item [C5]
The nozzle should either remain in the same position, move one step to
the right, left, top or bottom in every time instance.
Again this requires an implication such as if $\p{i}{j}{k} = 1$, then
one of $\p{i}{j}{k+1}, \p{i-1}{j}{k+1}, \p{i+1}{j}{k+1},
\p{i}{j-1}{k+1}$ or $\p{i}{j+1}{k+1}$ is $1$.
We encode this requirement using a linear constraint as follows:
$$
\p{i}{j}{k} \leq \p{i}{j}{k+1} + \p{i-1}{j}{k+1} + \p{i+1}{j}{k+1}
+ \p{i}{j-1}{k+1} + \p{i}{j+1}{k+1}
$$
Note that if $\p{i}{j}{k} = 0$, then the above inequality is trivially
true, hence, it does not impose any additional constraints.
On the other hand, if $\p{i}{j}{k} = 1$, this essentially states that
one of $\p{i}{j}{k+1}, \p{i-1}{j}{k+1}, \p{i+1}{j}{k+1},
\p{i}{j-1}{k+1}$ or $\p{i}{j+1}{k+1}$ is $1$.
Again, the constraint will change accordingly for the boundary cells, in that,
the terms corresponding to the indices that are out of the range will be eliminated.

\item [C6]
  The following constraint encodes the temperature constraints,
  namely, that the temperature should remain within $\T_l$ and
  $\T_u$ at all times and in all cells.

  \[\T_l \leq \t{i}{j}{k} \leq \T_u\]
\end{itemize}

The above will provide an annotated path that covers the print pattern
$P$, and satisfies the temperature evolution constraints and the
temperature bound constraints.
We need to provide an objective function such that minimization of the
same will provide the minimum cost valid annotated path.
Our broad idea is to capture the
index of the last printed cell  in the plan using the variable $m$.
We capture this by adding constraints $m \geq k$ for every $k$ where
printing happens in the path.
Note that printing happens at $k$ when $\sum_{(i, j) \in [m] \times
  [n]} \u{i}{j}{k} = 1$.
$$
m \geq k - 2d + 2d \sum_{(i, j) \in [m] \times
  [n]} \u{i}{j}{k}
$$
Note that if $\sum_{(i, j) \in [m] \times
  [n]} \u{i}{j}{k} = 1$, then the constraint reduces to $m \geq k - 2d
  + 2d = k$, as required.
On the other hand, if $\sum_{(i, j) \in [m] \times  [n]} \u{i}{j}{k} =
  0$, then the constraint reduces to $m \geq k - 2d$. Since $k \leq d$
  and $d > 0$, we have $k - 2d$ is a negative number, which is a
  trivial constraint that does not impose any restrictions.
  Note that $m$ satisfying these constraints only provides an upper
  bound on the last printed index in the plan, however, minimizing for
  $m$ not only ensures that $m$ coincides with the last printed index,
  but also that the plan itself is minimal.

\section{Illustrative Example}
In this section, we illustrate the encoding and the optimal path generation on the example in Figure \ref{fig:scenario}.
In the following,  we choose $\alpha =1$, $T_h = 1$, $\T_l=0$,
$\T_u=200$, time boundary $d=10$ and initial temperature of $\T(i, j)
= 75$ for each cell. 
We instantiate some of the constraints for this example to illustrate.
 
The constraint $C1$ for the center cell $(1,1)$ at time $1$ is as below. As seen from the following constraint, the temperature at center cell at time $1$ is related to temperature at that point at time $0$, the temperature of its four neighbor cells, and the extruder temperature.
\[\t{1}{1}{1} = \t{1}{1}{0} + \alpha( [\t{2}{1}{0} - \t{0}{1}{0}]/2\]
\[+ [\t{1}{2}{0} - \t{1}{0}{0}]/2) + \T_h \u{1}{1}{0}\]

The temperature constraints for a corner cell $(0,0)$ at time $1$, we obtain:
\[\t{0}{0}{1} = \t{0}{0}{0} + \alpha( [\t{1}{0}{0} - \t{0}{0}{0}]/2\]
\[+ [\t{0}{1}{0} - \t{0}{0}{0}]/2) + \T_h \u{0}{0}{0}\]

Constraint $C2$ captures that every cell in the print region is printed at some time, and every cell not in the print region is never printed.
The cosntraint for the center cell $(1,1)$, which corresponds to a print cell, that is, $P(1, 1) = 1$, is given by:
\[\u{1}{1}{0}+\u{1}{1}{1} +\u{1}{1}{2} +\cdots+ \u{1}{1}{10} = 1\]

And the constraint for the cell $(0,1)$ which is not in the print region is: 
\[\u{0}{1}{0}+\u{0}{1}{1} +\u{0}{1}{2} +\cdots+ \u{0}{1}{10} = 0\]

Constraint $C3$ states that if a cell is printed at time $k$, then the printed head is at that cell. For both the cells $(1,1)$ and $(0, 1)$, we have similar constraints.
\[\u{1}{1}{2} \leq \p{1}{1}{2}\]
\[\u{0}{1}{2} \leq \p{0}{1}{2}\]

The constraint $C4$ captures that at each time, the nozzle is exactly in one cell.
For time instance $2$, the constraint is given by:
\[\p{0}{0}{2}+\p{0}{1}{2}+\p{0}{2}{2}+\cdots+\p{2}{2}{2} = 1\]

The constraint $C5$ captures that the nozzle moves at most one step to one of its neighbors in one time unit elapse.
For example, if the nozzle is at cell $(1,1)$ at time $1$, it can move $1$ step to the left or right of the cell horizontally or vertically or remain at the cell $(1,1)$ at time $2$.
This is captured by:
\[\p{1}{1}{1} \leq \p{1}{1}{2} + \p{0}{1}{2} + \p{2}{1}{2}+\\
\p{1}{0}{2} + \p{1}{2}{2}\]

But for corner cell (0,0) at time 1, the next movement can be one step to the right side of the cell horizontally or down side of the cell vertically.

\[\p{0}{0}{1} \leq \p{0}{0}{2}, + \p{1}{0}{2} + \p{0}{1}{2}\]

The constraint $C6$ captures that the temperature of each cell should be between $\T_l$ and $\T_u$, which in this example for cell $(1,1)$ at time $1$, with $\T_l=0$, and $\T_u=200$, is given by:
\[0 \leq \t{1}{1}{1}\leq 200 \]

Finally, our goal is to minimize $m$ corresponding to the index of the last printed cell. For time $1$, the constraint on $m$ is given by:
\[m \geq 1 - 2\times 10 + 2\times 10 (\u{0}{0}{1}+\u{0}{1}{1}+\u{0}{2}{1}+\cdots+\u{2}{2}{1})\]
  
\begin{figure}[thpb]
      \centering

      \includegraphics[scale=.3]{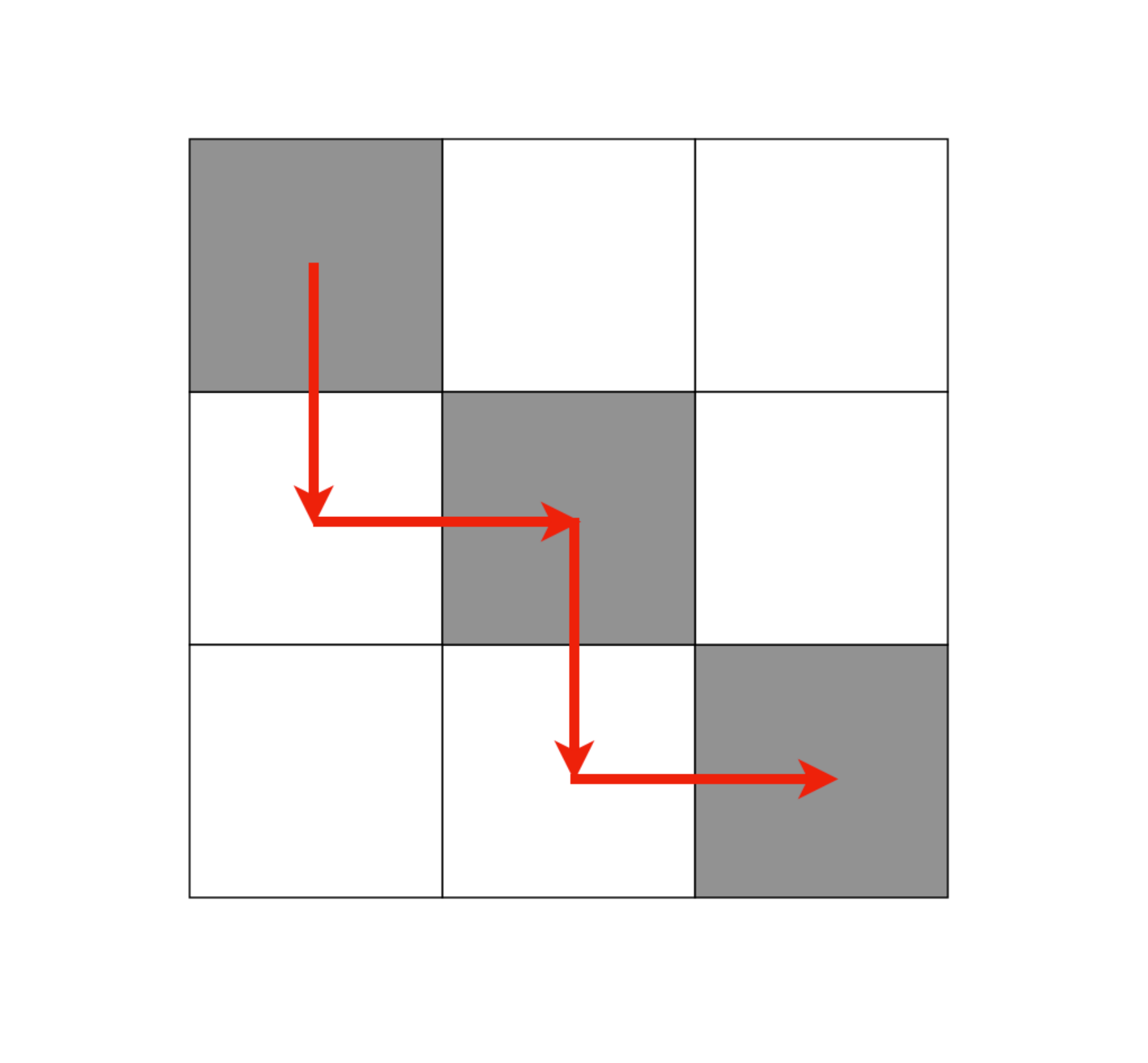}
      \caption{The Gurobi's path optimization on $3 \times 3$ bed with gray print region}
      \label{fig:plan}
    \end{figure}

After generating all the constraints, we sent the constraints to Gurobi solver and Figure \ref{fig:plan} shows the obtained path which is shown by red arrows.

The value of returned $m$ was $4$, and printer was printing at time instants $0$, $2$ and $4$, which corresponds to the gray cells.
    When we decreased the temperature range to $\T_l = 65$ and $\T_u = 85$, we obtained a larger $m$ and a longer path where the printer was sitting idle at the same cell for more than $1$ time unit. We believe that this is to ensure that the temperature constraints are satisfied, since, as time progresses, the temperature distributes and reduces to be within the range.
\section{Experimental Evaluation}
In this section, we briefly describe our implementation and
experimental evaluation.
We implemented a Python tool box to automatically generate constraints
corresponding to $\enc(\S, d)$ in the Gurobi input format, given a
scenario $\S$ and a time bound $d$.
Gurobi is an MILP solver, which returns the optimal assignment.
We have implemented an algorithm to extract the optimal annotated plan
from the optimal assignment.

We consider a diagonal pattern, that is, the print region
corresponds to the diagonal cells in an $n \times n$ grid to evaluate our method.
We experiment with varying grid sizes (increasing values of $n$),  and
with two different settings for temperature bounds.
The results are presented in Table \ref{tab:one} and Table \ref{tab:two}.
Our experiments have been conducted on 2.8 GHz Intel Core i7 Laptop. 
Dimensions represents the size of the grid,
$d$ denotes the time bound,
$m$ denotes cost of the returned path (length of the path until the
last printed cell),
the time taken to encode the constraints in seconds and
the time taken by Gurobi to find the value of $m$ (and the optimal assignment).
     
\begin{table}[h]
\caption{Diagonal print pattern with $\T_l = 0$ and $\T_u = 200$}
\label{tab:one}
\begin{center}
\begin{tabular}{|c|c|c|c|c|}
\hline
Dimensions & 2$\times$ 2 & 3$\times$ 3 & 5$\times$ 5 & 7$\times$ 7\\
\hline
d & 10 & 10 & 10 & 15\\
\hline
m  & 2 & 4 & 8 & 14\\
\hline
Encoding time & 0.035 & 0.06 & 0.194 & 0.581\\ 
\hline
MILP solving time & 0.065 & 0.12 & 0.848 & 10.306\\
\hline
\end{tabular}
\end{center}
\end{table}

First, we observe from Table \ref{tab:one} for the diagonal printing
pattern with $\T_l = 0$ and $\T_u = 200$ that both the encoding time
and the MILP solving time increase as we increase the size of the grid.
However, MILP solving time, increases much faster than the encoding
time, which points to the fact that MILP is an NP-complete problem.
Also, we had a time-out of $1$ minute, and the procedure times out
for $9 \times 9$ grid.
In all the cases, we obtained the shortest path to be the step plan
along the diagonal as illustrated in Figure \ref{fig:plan}.

\begin{table}[h]
\caption{Diagonal print pattern with $\T_l = 65$ and $\T_u = 85$}
\label{tab:two}
\begin{center}
\begin{tabular}{|c|c|c|c|c|}
\hline
Dimentions & 2$\times$ 2 & 3$\times$ 3 & 5$\times$ 5 & 7$\times$ 7\\
\hline
d & 10 & 10 & 10 & 15\\
\hline
m  & 6 & 7 & 9 & NF\\
\hline
Encoding time & 0.039 & 0.072 & 0.19 & NF\\ 
\hline
MILP solving time & 0.17 & 0.12 & 0.62 & NF\\
\hline
\end{tabular}
\end{center}
\end{table}

Table \ref{tab:two} for the diagonal printing
pattern with $\T_l = 65$ and $\T_u = 86$, we have a similar
observation about the encoding and MILP solving time.
However, with this reduced range for the temperature, we observe that
the values of $m$ are relatively larger.
In fact, when given a bound of $15$ for the $7 \times 7$ grid, the
procedure returned infeasible, indicating a potential plan that
satisfies all the constraints would be longer.
We observed that the path consisted sometime of repeated consecutive
positioning of the  nozzle at a particular cell.
We conjecture that this is to satisfy the temperature constraints, and
may corresponding to allow ``cooling'' of the cell temperatures.

\section{Conclusion}
In this paper, we addressed the problem of optimal path planning with
spatio-temporal and thermal constraints, that arise naturally in the
context of 3D printer tool path planning for superior structural properties.
We considered a discrete version of the problem, and reduced the
problem to MILP problem, which we solved using off-the-shelf solvers.
Our experimental analysis demonstrated the feasibility of the
approach.
In the future, we will explore a continuous version of the problem,
where we consider a PDE heat equation and generate continuous plans
that specify velocity and acceleration of the print head.
\section*{Acknowledgments}
Pavithra Prabhakar was partially supported by NSF CAREER Award No. 1552668 and  ONR YIP Award No. N000141712577.

\bibliographystyle{IEEEtran}
\bibliography{IEEEabrv,references}
\end{document}